\documentclass{Interspeech}

\interspeechcameraready

\title{SpeechLLM: Unified Speech and Language Model for Enhanced Multi-Task Understanding in Low Resource Settings}

\author[affiliation={1}]{Jaekwon}{Yoo}
\author[affiliation={1}]{Kunal}{Chandiramani}
\author[affiliation={2}{*}]{Divya}{Tadimeti}
\author[affiliation={1}]{Abenezer}{Girma}
\author[affiliation={1}]{Chandra}{Dhir}

\affiliation{}{JPMorganChase}{USA}
\affiliation{}{Columbia University}{USA}
\email{\{jaekwon.yoo, kunal.chandiramani, abenezer.girma, chandra.dhir\}@jpmchase.com, divya.tadimeti@columbia.edu}

\usepackage{comment}
\usepackage{soul}
\usepackage{multirow}
\usepackage{booktabs}
\usepackage{graphicx}

\begin{document}

\maketitle
\let\thefootnote\relax\footnote{* Work was done at JPMorganChase.}

\begin{abstract}
While integrating speech encoder with LLM requires substantial data and resources, use cases face limitations due to insufficient availability. To address this, we propose a solution with a parameter-efficient adapter that converts speech embeddings into LLM-compatible tokens, focusing on end-to-end automatic speech recognition (ASR), named entity recognition (NER), and sentiment analysis (SA). To reduce labeling costs, we employ an LLM-based synthetic dataset annotation technique. The proposed adapter, using $7\times$ fewer trainable parameters, achieves significant performance gains: a 26\% relative Word Error Rates(WER) improvement on the LibriSpeech ASR task, a 6.3\% relative F1 score increase on the NER task, and a 32\% relative F1 score boost on the SA task. Moreover, using advanced techniques such as adding a classifier regularizer and optimizing the LLM with Low-Rank Adaptation(LoRA) yields notable performance gains, with Spoken Language Understanding Evaluation(SLUE) score improvement of 6.6\% and 9.5\%.
\end{abstract}

\section{Introduction}
Recent advancements in speech and language tasks have shown a significant shift in interest from domain-specific models to those trained on much larger and more general corpora. In particular, Large Language Models (LLMs) have demonstrated a remarkable ability to perform a wide variety of tasks and show great promise for more general language understanding~\cite{achiam2023gpt}. The immense size of the training corpora for these models means that they are exposed to a vast array of words, some of which may occur only rarely but are quite important for certain use cases like Named Entity Recognition (NER). The integration of ASR with LLMs presents a promising approach to overcoming the limitations inherent in each technology. Specifically, this combination addresses the constraint of relatively smaller text datasets typically used in automatic speech recognition (ASR) and compensates for the inability of LLMs to process audio data directly \cite{peng2025surveyspeechlargelanguage, radford2023robust, zhang2024tinyllama, ma2024embarrassingly}.

Despite a recent upswing in efforts~\cite{zhang2023speechgptempoweringlargelanguage,tang2023salmonn,chu2023qwen}, the best strategies for combining large generic speech and large generic language models into a single end-to-end (E2E) model are not well understood. Even simple methods have shown good promise~\cite{ma2024embarrassingly}, but there could be many other strategies that have not yet been tried. Attention has been paid to various adapter and joint training strategies, as well as big multi-modal models trained from scratch~\cite{zhan2024anygpt}.

Tasks that might particularly benefit from an E2E approach incorporating both generic speech and language models are NER and Sentiment Analysis (SA). Traditionally, NER and SA uses a pipeline approach, where NER and SA models are applied to transcripts produced by ASR systems. However, this method may suffer from error accumulation, where transcription errors are propagated to downstream NLP systems, thereby reducing NER and SA performance. Similarly, adapting ASR models to perform NER may struggle with rare terms and lack of enough training data, as the size of the models and data for ASR systems is still orders of magnitude smaller than that of LLM systems. To address the shortage of training data, some studies, such as \cite{yu2025ve}, have proposed synthetic NER data generation methods.

Recent research has focused on integrating speech models into a broad array of NLP tasks beyond simple transcription \cite{meeus2023whisper, arora2024evaluation, arora2024universlu, li2024prompting, chen2024salm}. For instance, spoken language understanding (SLU) models have been used to perform tasks like NER and SA directly from speech, removing the need for a pipeline approach \cite{li2024prompting, ayache2024whisperner, yang2023chinese, bogdanov2024nunerentityrecognitionencoder}. However, these methods often require training either the ASR encoder or the LLM's decoder, resulting in significant training costs, or they depend on zero-shot approaches, which results in poor performance. In \cite{li2024using}, the concept of integrating a speech encoder with an LLM was applied to Chinese ASR and NER tasks. Another study proposed WavLLM \cite{hu2024wavllm}, which combines Whisper and WavLM encoders with LLaMA-2-chat. However, this approach involved multiple speech encoders and adapters to transform the embedding from the speech encoder to LLM, which significantly increase the complexity and computational requirements. 

In this paper, we present SENSE-ASR: Speech Encoder Network for Sentiment Analysis and Entity recognition, a model that seamlessly integrates a speech encoder with Large Language Models by training the adapter layer while keeping the speech encoder and LLM either frozen or unfrozen. This setup allows an E2E optimization, efficiently handling multiple speech-related tasks. We validate our method by simultaneously performing transcription, NER and SA tasks and demonstrating that SENSE-ASR achieves benchmark performance on ASR, NER and SA tasks with minimal training parameters. We selected the NER and SA tasks for our experiments because it challenges the LLM to not only map speech features to text tokens but also to comprehend the semantic meaning of the ASR transcription.

The main contributions of this paper are 
\begin{itemize}
    \item Developed a lightweight adapter for an E2E ASR-LLM integration, enabling multi-tasking of ASR/NER and ASR/SA.  
    \item Implemented a pre-training NER data annotation methodology using LLM, leveraging a smaller subset of human-annotated NER data.
    \item Developed a regularization classifier to enhance NER and SA
    \item Achieved new E2E model benchmarks for NER on the SLUE-VoxPopuli dataset and for SA on the SLUE-VoxCeleb dataset.
\end{itemize}

\section{Model Architecture}
\subsection{Overview of SpeechLLM}

\begin{figure}[htbp]
\centerline{\includegraphics[width=0.5\textwidth]{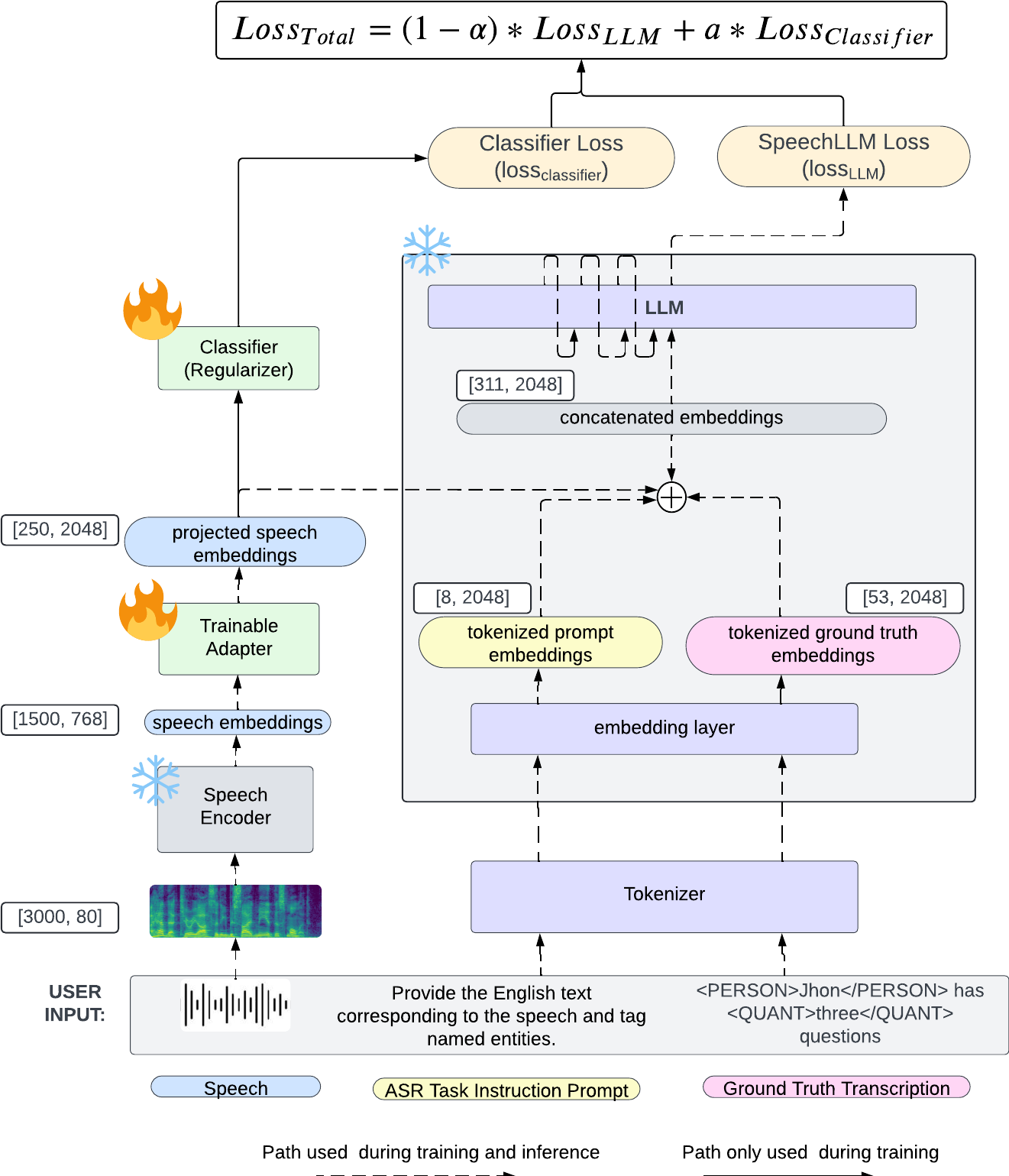}}
\caption{SpeechLLM Architecture.}
\label{fig:speechllm}
\end{figure}

The proposed SENSE-ASR architecture, as shown in Fig.\ref{fig:speechllm}, integrates a frozen Whisper speech encoder with a pre-trained TinyLlama language model using a simple trainable adapter.
The training input to the SENSE-ASR consists of three components: speech, task instruction prompt, and the ground truth text.

Given the input speech, its Mel-spectrogram is obtained and passed into the Whisper encoder. The Mel-spectrogram has $80$ mel frequency bands and $3000$ time steps, corresponding to $30$ seconds of speech, which is padded or trimmed as necessary meeting the speech-encoder input requirements. The speech-encoder processes the Mel-spectrogram and outputs a $1500\times768$ dimensional feature embeddings.

The embeddings are then passed into the proposed trainable adapter, which projects the embeddings to a $2048$-dimensional vector, making the speech embeddings compatible with the LLM's input embeddings. The output of the adapter is also fed into the classifier for regularization purposes.

Simultaneously, the input prompt and the ground truth (during training only) are tokenized and passed through the LLM's embedding layer, resulting in embeddings with a 2048 feature dimension. The embeddings for all three input components (speech embedding post-adapter, prompt and ground truth) are concatenated along the sequence length dimension. These concatenated embeddings are then passed into the LLM for auto-regressive generation of the output tokens.

During training, the input format is ( \textit{\texttt{USER: <S> <P> ASSISTANT: <GT>}}), where \texttt{<S>} is the speech, \texttt{<P>} is the ASR/NER/SA task instruction prompt, and \texttt{<GT>} is the ground truth text. During inference, the input format is: (\textit{\texttt{USER: <S> <P> ASSISTANT:}}), where the model generates the transcription following the prompt.

In the following section, we provide a detailed discussion of each component individually.

\subsection{Speech Encoder}

The Whisper ASR family of models~\cite{radford2023robust} are transformer-based models trained on transcribed speech data. To integrate this model with LLMs and create a multi-modal model, we utilize only the Whisper encoder, as it generates effective representations of the speech signal that can be consumed by the LLM. We experimented with both small and medium-sized pre-trained Whisper encoder.

\subsection{Large Language Model}
TinyLlama is trained on a massive dataset comprising a cumulative total of 3 trillion tokens \cite{zhang2024tinyllama}. The vast size of training data means this model has likely encountered a much larger fraction of potential entities than any ASR model by itself. By integrating this model with the Whisper encoder, we expect to be able to analyze sentiment better and identify rare entities more accurately.

\subsection{Adapter}
The proposed adapter module in SENSE-ASR serves as a bridge between the speech encoder and the LLM, transforming speech embeddings into a format compatible with the LLM's embedding space. This module comprises two main components: down-sampling and projecting.

\subsubsection{Down-sampling}

The down-sampling component is designed to reduce the size of the input embeddings from the speech encoder. This is achieved through the following steps:

\begin{itemize}
    \item Adaptive Average Pooling: The input embeddings from the Whisper encoder are passed through an adaptive average pooling layer. This layer reduces the size of the embeddings to a specified pooling size, which we set to 250. Adaptive average pooling ensures that the output size is consistent, regardless of the input size, by averaging the values within each pooling window.
    \item Layer Normalization: After pooling, the embeddings are normalized using layer normalization. This step ensures that the pooled embeddings have a consistent scale and distribution.
\end{itemize}

\subsubsection{Projection}

The projecting component is responsible for transforming the normalized embeddings into a dimensionality that is compatible with the LLM. This is done through a linear projection layer:

\begin{itemize}
    \item Linear Projection: The normalized embeddings are passed through a linear layer that projects them to the reduced output dimensionality required by TinyLlama. This linear transformation adjusts the feature dimensions of the embeddings, making them suitable for input into the LLM.
\end{itemize}

By combining these two components, the adapter module effectively downsamples and projects the speech embeddings, ensuring they are in the correct format for the LLM. This integration allows SENSE-ASR to process both speech and text inputs.

\subsubsection{Classifier (Regularizer)} \label{sec:NER-classifier}
The regularizer module is designed to optimize the trainable adapter for improved NER or SA predictions. The classifier utilizes a three-layer 2D convolutional neural network. The output from the final convolutional layer is passed to a fully connected neural network, which maps the extracted features to entity classes for the NER task and sentiment classes for the SA task within the dataset. The predicted entity and sentiment classes are then compared to the ground truth classes obtained from the labeled NER or SA transcripts.

The classifier loss is calculated based on the specific task, either NER or SA. For NER, a weighted Binary Cross-Entropy loss is employed to address class imbalance within the dataset. This approach assigns higher weights to underrepresented classes and lower weights to more prevalent ones, ensuring balanced loss calculation across all entities. For SA, the loss is calculated using Cross-Entropy. Finally, the classifier's loss ($\mathbf{Loss_{classifier}}$) is integrated with the LLM's cross-entropy loss ($\mathbf{Loss_{LLM}}$), serving as a regularizer, as illustrated in the equation below:

\begin{equation}
    \mathbf{Loss_{total} = (1-\alpha)*Loss_{LLM} + \alpha*Loss_{classifier}}
\end{equation}
where $\alpha$ is a hyperparameter used to control the influence of the regularizer during the training. Once training is complete, the classifier will not be used for inference.

\subsection{Multi-tasks \& Metrics}
\label{sec:metrics}
\begin{itemize}
    \item \textbf{Speech Transcription}: The ASR task involves converting spoken language utterance into written texts. This task is evaluated using the standard Word Error Rate (WER) metrics.
    \item \textbf{Named Entity Recognition} (NER): This task involves identifying named entities and their corresponding tags in spoken language. It is evaluated using two metrics: the micro-averaged F1 score, also referred as F1 score in the remainder of the paper, and the label-F1 score. The F1 score assesses the accuracy of predicted named entity phrase and tag pairs for each sentence. In contrast, the label-F1 score focuses solely on the accuracy of tag predictions, evaluating the model's ability to correctly identify and classify entity tag labels. For instance, if the ground truth is \textless PERSON\textgreater John\textless /PERSON\textgreater and the prediction is \textless PERSON\textgreater Jon\textless /PERSON\textgreater, the F1 score would mark this as incorrect due to the entity transcription error (\textit{Jon} $\neq$ \textit{Jhon}). However, the label F1 score would consider it correct, as it only evaluates the accuracy of the entity label prediction (\textit{PERSON}), regardless of the entity transcription error.
    We map the NER tag annotations into special symbols for a effective and efficient representation as described in \cite{yadav2020endtoendnamedentityrecognition, ghannay2018end}.

    \item \textbf{Sentiment Analysis} (SA) : This task focuses on identifying the sentiment of speakers, as Positive, Negative, and Neutral classes. The macro-averaged F1 score is used to asses SA task.
    \item \textbf{SLUE Score}: The SLUE score as discussed in \cite{shon2022slue} is the average of the primary metrics for the three tasks (ASR, NER and SA) defined as: $SLUE_{score} =$ $\textstyle \frac{1}{3} \left( \left( 100 - \frac{\text{WER}_{\text{ASR-VP}} + \text{WER}_{\text{ASR-VC}}}{2} \right) + \text{F1}_{\text{NER-VP}} + \text{F1}_{\text{SA-VC}} \right)$.
\end{itemize}

\subsection{Dataset}
\begin{sloppypar}
We conducted our experiments using the benchmark ASR dataset LibriSpeech\cite{panayotov2015librispeech} along with spoken language understanding dataset; SLUE-VoxPopuli(VP) dataset for NER task and SLUE-VoxCeleb(VC) dataset for SA task \cite{shon2022slue, wang2021voxpopuli, nagrani2017voxceleb}. 
\end{sloppypar}

\subsubsection{Synthetic data generation}
\label{sec:LibriSpeech-NER}
To address the limitation of having only 15 hours of NER training data in the SLUE-VoxPopuli dataset, we devised a pre-training strategy. This approach leverages the human-annotated SLUE-VoxPopuli data to synthetically annotate 960 hours of the Librispeech ASR dataset with entities. To achieve this, we employed a systematic prompt engineering approach, evaluating various LLM prompts on SLUE-VoxPopuli dev dataset. For each prompt, we use the LLM to annotate the SLUE-VoxPopuli dev dataset and then assessed the LLM-annotated NER data against the human-annotated SLUE-VoxPopuli data using the F1 score.

We started our prompt engineering experiments with a specific prompt proposed by NuNER \cite{bogdanov2024nunerentityrecognitionencoder} paper while restricting it to identify entities only within $7$ categories to match OntoNotes Release $5.0$ \cite{hovy2006ontonotes} guidelines and SLUE-VoxPopuli entity labels. However, the zero shot prompt in \cite{bogdanov2024nunerentityrecognitionencoder} resulted in a poor F1 score of 0.3. To improve performance, we tried few shot prompting, taking inspiration from \cite{ashok2023promptner} by adding $1000$ SLUE-VoxPopuli train-examples (balanced over the distribution of entity tags) as part of the prompt, boosting the NER F1 score to $0.77$. This score is quite competitive with the estimated human performance of $0.79$ F1 score as reported in the SLUE paper \cite{shon2022slue}. 

We use GPT-4o to annotate $960$ hours of Librispeech transcription data with NER labels. We then apply two filtration steps: one to detect and remove sentences with possible LLM hallucination, and second LLM re-verification step by passing all unique entities and their corresponding tags into the LLM to identity meaningless entities to be removed from the dataset. Evaluation on SLUE-VoxPopuli dev-set shows these post-processing steps helped in improving the F1 score by 2\% absolute. 

The detailed distribution of entity tags over train, dev and test sets is shown in Table 1.

\begin{table}[h]
\centering
\resizebox{0.47\textwidth}{!}{%
\begin{tabular}{l|l|l|l|l|l|l|l}
\hline
\textbf{} & \textbf{PERSON} & \textbf{PLACE} & \textbf{WHEN} & \textbf{NORP} & \textbf{ORG} & \textbf{QUANT} & \textbf{LAW} \\ \hline \hline
\textbf{Train} & 122526 & 37283 & 16846 & 13254 & 2981 & 2775 & 2104 \\
\textbf{Dev} & 1668 & 369 & 213 & 115 & 30 & 146 & 8 \\
\textbf{Test} & 1454 & 354 & 186 & 176 & 22 & 328 & 12 \\ \hline
\end{tabular}
}
\caption{Distribution of synthetically generated NER labels for LibriSpeech. Refer to section \ref{sec:LibriSpeech-NER}}
\label{table:finetuning_configurations}
\end{table}

\subsection{Training Strategies}
\label{sec:training}

The SENSE-ASR model includes two configurations: Whisper small and Whisper medium as speech encoders \cite{radford2023robust}, both paired with TinyLLama \cite{zhang2024tinyllama} as a decoder, linked through the proposed adapter layer. As illustrated in Table~\ref{tab:SLUE-VoxPopuli}, we experiment with three types of data training strategies:
\begin{sloppypar}
\begin{itemize}
    \item \textit{Random:} This method involves training the SENSE-ASR adapter directly on the human-annotated SLUE fine-tune split dataset.
    \item \textit{LibriSpeech-ASR (LS-ASR):} This strategy involves pretraining the SENSE-ASR adapter on LibriSpeech ASR data, followed by fine-tuning directly on human-annotated SLUE dataset.
    \item \textit{LibriSpeech-ASR-NER (LS-ASR+NER):} In this approach, the SENSE-ASR adapter is first pre-trained on LibriSpeech data for ASR task, followed by training for ASR and NER task on a synthetically annotated LibriSpeech NER dataset, and concludes with fine-tuning on the human-annotated SLUE-Voxpopuli dataset.
\end{itemize}
\end{sloppypar}
During training, data augmentation techniques such as speed perturbation and SpecAugment \cite{Park_2019} are applied.

\section{Experimental Results and Discussion}

In this section we evaluate the proposed SENSE-ASR model using the Librispeech, SLUE-VoxPopuli(VP) and SLUE-VoxCeleb(VC) dataset against the baselines presented in \cite{ma2024embarrassingly, shon2022slue} using metrics discussed in \ref{sec:metrics}. The ASR baseline benchmarks on the Librispeech dataset are derived from  the study in \cite{ma2024embarrassingly}, which is similar to ours but focuses solely on ASR. Meanwhile, the NER and SA baseline benchmarks are derived from the study by \cite{shon2022slue}, which incorporates a linear layer on top of the W2V2 \cite{baevski2020wav2vec} and HuBERT \cite{hsu2021hubert} pre-trained models and fine tunes them on the SLUE-VoxPopuli and SLUE-VoxCeleb dataset. During inference time, the results for the proposed method were obtained using a beam size of 5, a temperature of 1.0, a repetition penalty of 2.0, and a length penalty of 0.5. When training with SA and NER Classifier we use an alpha of 0.2.

\subsection{LibriSpeech}\label{sec:LibriSpeech-exp}
The first phase of training involves validating SENSE-ASR on its speech-to-text ASR task, trained solely on the $960$ hours of LibriSpeech data. The model is trained for $20$ epochs with a learning rate of $10^{-4}$, a batch size of $4$, and a linear decay learning rate scheduler with $1000$ warm-up steps. The optimal adapter is selected by evaluating each epoch's adapter on the librispeech dev-clean set and selecting the best performing one based on the WER score. Since the Librispeech dataset contains only human-labeled transcripts, we assess the proposed approach for its ASR capability.

\begin{table}[h]
  \centering
  \resizebox{0.47\textwidth}{!}{%
  \begin{tabular}{lccc}
    \toprule
    Speech Encoder & \# Parameters (M) & test-clean & test-other \\
    \midrule
    \multicolumn{4}{l}{\textbf{Whisper Small}} \\
    SLAM-ASR benchmark\cite{ma2024embarrassingly} & 13.1 & 5.94 & 11.5  \\
    Proposed Adapter & 1.6 & 5.22 & 10.36 \\
    \midrule
    \addlinespace
    \multicolumn{4}{l}{\textbf{Whisper Medium}} \\
    SLAM-ASR benchmark\cite{ma2024embarrassingly} & 15.2 & 5.01 & 8.67  \\
    Proposed Adapter & 2.1 & 3.67 & 8.30 \\
    \bottomrule
  \end{tabular}
  }
  \caption{The proposed SENSE-ASR WER (\%) performance on LibriSpeech dataset.}
  \label{tab:librispeech-asr}
\end{table}

As shown in Table~\ref{tab:librispeech-asr}, compared to the baseline \cite{ma2024embarrassingly}, our proposed adapter with whisper medium achieved a 26\% and 4\% improvement in WER on the test-clean and test-other sets, respectively, while using only 15\% of the trainable parameters. 

We compared our proposed adapter's performance in ASR experiments with the SLAM-ASR benchmark \cite{ma2024embarrassingly}, as other studies, such as those in \cite{chu2023qwen, tang2023salmonn, hu2024wavllm}, use larger speech encoder and 7 billion parameter LLM models, making them unsuitable for direct comparison. Furthermore, to demonstrate the effectiveness of the proposed model's capabilities in spoken language understanding tasks, we extended our analysis to evaluate its performance on NER and SA tasks using the SLUE datasets. 

\subsection{SLUE-VoxPopuli data} \label{slue_voxpoplu_data}
The SLUE-VoxPopuli dataset provides both human-labeled transcripts and NER tags, allowing us to evaluate the proposed approach for the multi-task objectives of ASR and NER. We begin our NER training with adapters derived from ASR training in Sec. \ref{sec:LibriSpeech-exp} , referred to as LS-ASR in Table \ref{tab:SLUE-VoxPopuli}. We then pre-train the SENSE-ASR model with 960 hours of synthetic labeled Librispeech-NER data. This NER pre-training is conducted for 10 epochs with a learning rate of $10^{-4}$, batch size of 48, and a linear decay warm-up scheduler with 1000 warm-up steps. We refer to this pre-trained adapter as LS-ASR+NER in Table \ref{tab:SLUE-VoxPopuli}. The optimal pre-trained LibriSpeech-ASR-NER adapter is selected by evaluating each epoch's adapter on LibriSpeech-NER dev data and choosing the adapter with the best F1-Score. 

The final phase of training includes training on 15 hours of SLUE-VoxPopuli dataset, both with and without a NER-Classifier as a loss regularizer, as mentioned in section \ref{sec:NER-classifier}. We conduct this training for 200 epochs with a learning rate of $10^{-3}$, a batch size of 16, and a linear decay scheduler with 2000 warm-up steps. When training with the NER-Classifier, we set the pos\_weight for each class as the total number of tags divided by the number of tags for that class.

\begin{table}[h!]
  \centering
  \resizebox{0.47\textwidth}{!}{
  \begin{tabular}{llcccc}
    \toprule
    Speech Model & Pretrained Data & WER(\%) & F1(\%) & Label F1(\%)  \\ 
    \midrule
    \multicolumn{5}{l}{\textbf{SLUE Benchmark\cite{shon2022slue}}} \\
    W2V2-B-LS960 + LM$^{**}$ & - & 12.3 & 63.4 & 71.7  \\ 
    W2V2-L-LL60K + LM$^{**}$ & - & \textbf{9.3} & 64.8 & 73.4 \\
    \midrule
    \multicolumn{5}{l}{\textbf{Proposed E2E Approach}} \\
    \multirow{3}{*}{Whisper Small} 
    & Random  & 19.3 & 54.2 & 70.0  \\
    & LS - ASR & 15.0 & 59.0 & 74.4  \\
    & LS - ASR+NER & 14.7 & 59.3 & 73.9  \\
    \midrule
    \multirow{3}{*}{Whisper Medium} 
    & Random  & 14.7 & 62.0 & 73.7  \\
    & LS - ASR & 11.6  & 64.3 & 78.7  \\
    & LS - ASR+NER & 11.4 & 65.0 & 79  \\
    \midrule
    \multirow{3}{*}{}
    + NER Classifier & LS - ASR+NER & 10.9 & 65.8 & 78.8  \\
    + LLM LoRA fintuning & & 10.6 & \textbf{68.9} & \textbf{81.2} \\
    \bottomrule
  \end{tabular}
  }
  \caption{Performance of the SENSE-ASR models on SLUE-VoxPopuli. LS refers to the LibriSpeech Dataset. Refer to section \ref{sec:training} for more details. **TED LM with Beam decoding of 500.}
  \label{tab:SLUE-VoxPopuli}
\end{table}

We evaluated the SENSE-ASR on the SLUE-VoxPopuli NER test dataset, as shown in Table~\ref{tab:SLUE-VoxPopuli}. With random initialization of adapter weights and training only on the SLUE-VoxPopuli ASR+NER dataset, the Whisper Medium variant achieved an F1 score of $61.5\%$. However, pre-training the adapter with LS-ASR data on the ASR task alone, followed by fine-tuning on the SLUE-VoxPopuli ASR+NER task, led to a $4.55\%$ relative improvement in F1 score.  Further improvement was achieved by training on synthetically annotated LibriSpeech-ASR+NER data before fine-tuning on SLUE-VoxPopuli, yielding an additional $1.09\%$ improvement. Overall, we achieved a $5.69\%$ relative improvement in F1 score compared to a random initialization. These clearly demonstrates the advantages of the proposed training strategy and highlight the effectiveness of the adapter in continuously learning and enhancing performance.

We included NER classification loss as a regularizer, as discussed in Sec. \ref{sec:NER-classifier}, in the training of SENSE-ASR adapter. Experimental results show $6.81\%$ relative improvement over a random initialization. Furthermore, the LLM was fine-tuned using LoRA, with $\gamma=32$ and $\alpha=32$ applied across all query, key, and value matrices of the self-attention mechanism, resulting in 0.58\% of the total parameters. This fine-tuning achieved an F1 score of 68.1\% and a Label F1 score of 81.1\%.

Upon applying the LLM beam decoding of 5 mentioned in \ref{sec:LibriSpeech-exp}, we achieved an improved and SOTA F1-score of 68.9\%. Using LS960, we show an relative improvement of 6.8\% and 10.6\% in F1 and Label F1 score compared to the SLUE benchmark, respectively, over W2V2-B-LS960 + LM. This also resulted in a SLUE score of 79.2 compared against the benchmarking of 77.8 in \cite{shon2022slue}.

\subsection{SLUE-VoxCeleb data}
The SLUE-VoxCeleb dataset provides both human-labeled transcripts and Sentiment Analysis classes, allowing us to evaluate the proposed approach for the multi-task objectives of ASR and SA. The dataset categorizes sentiment into five classes: Positive (indicating happiness and a positive attitude), Negative (indicating negative emotions), Neutral (indicating no emotion), Mixed (a combination of positive and negative emotions), and Disagreement (indicating conflicts in annotation). In line with study that annotated the data \cite{nagrani2017voxceleb}, we did our experiments and evaluations on the Positive, Negative, and Neutral classes. 

We begin our SA training with random and adapters derived from ASR training in Sec. \ref{sec:LibriSpeech-exp} , referred to as LS-ASR in Table \ref{tab:SLUE-VoxPopuli}. We utilized 12.8 hours of training data and 3.2 hours of development data for training and tuning tasks, and assessed the model's performance on a 7.8-hour test split, as reported in this paper. We conduct the SA training for 50 epochs with a learning rate $5*10^{-4}$, batch size 6, and a linear decay scheduler with 3000 warm-up steps. 

\begin{table}[h!]
  \centering
  \resizebox{0.41\textwidth}{!}{
      \begin{tabular}{llcc}
        \toprule
        Speech Model & Pretrained Data & WER(\%) & F1(\%)  \\ 
        \midrule
        \multicolumn{4}{l}{\textbf{SLUE Benchmark\cite{shon2022slue}}} \\
        W2V2-B-LS960 + LM$^{**}$ & - & 16.1 & 48.1 \\ 
        W2V2-L-LL60K + LM$^{**}$ & - & \textbf{11.1} & 49.8  \\
        \midrule
        \multicolumn{4}{l}{\textbf{Proposed E2E Approach}} \\
        \multirow{3}{*}{Whisper Small} 
        & Random  & 19.9 & 54.4  \\
        & LS - ASR & 16.0 & 60.9  \\
        \midrule
        \multirow{3}{*}{Whisper Medium} 
        & Random  &  16.7 & 58.9 \\
        & LS - ASR & 13.1  & 61.0  \\
        %\midrule
        \multirow{3}{*}{}
        + SA Classifier & LS - ASR & 12.3 & 63.6  \\
        + LLM LoRA fintuning & & 11.5 & \textbf{65.9} \\
        \bottomrule
      \end{tabular}
  }
  \caption{Performance of the SENSE-ASR models on SLUE-VoxCeleb Sentiment Analysis task. LS refers to the LibriSpeech Dataset. Refer to section \ref{sec:training} for more details. (**TED LM with Beam decoding of 500 for ASR task.)}
  \label{tab:SLUE-VoxCeleb}
\end{table}

The results presented in Table \ref{tab:SLUE-VoxCeleb} highlight the performance of various speech models on the SLUE-VoxCeleb Sentiment Analysis task, comparing both the SLUE Benchmark models and the proposed E2E approach. The SLUE Benchmark models, specifically W2V2-B-LS960 + LM and W2V2-L-LL60K + LM, achieved Word Error Rates (WER) of 16.1\% and 11.1\%, respectively, with corresponding F1 scores of 48.1\% and 49.8\%. These benchmarks serve as a reference point for evaluating the effectiveness of the proposed models. 

In comparison, the proposed E2E approach, utilizing Whisper Small and Whisper Medium models, demonstrated notable improvements in F1 scores, particularly when pre-trained with LS-ASR data and enhanced with additional techniques. The Whisper Small model, when trained with LS-ASR, achieved a WER of 18.0\% and an F1 score of 60.9\%, while the Whisper Medium model further improved these metrics to a WER of 13.1\% and an F1 score of 61.0\%. The integration of the SA Classifier and LLM LoRA fine-tuning further enhanced the performance further, with the latter achieving an F1 score of 65.8\% and a WER of 12.9\%. Although our proposed models see slightly higher WER scores compared to the benchmark, this is attributed to the benchmark's use of TED Language Models, which leverage out-of-domain trigram models with a large beam decoding size of 500. Despite this, the proposed E2E approach excels in sentiment analysis accuracy, as evidenced by the higher F1 scores, showcasing the potential of ASR training, SA Classifier, and LLM LoRA fine-tuning in advancing sentiment analysis capabilities in speech models.

\begin{table}[h!]
  \centering
  \resizebox{0.47\textwidth}{!}{
  \begin{tabular}{l@{\hskip 0.1in}l@{\hskip 0.1in}c@{\hskip 0.1in}c@{\hskip 0.1in}c@{\hskip 0.1in}c}
    \toprule
    Speech Model & SLUE & \multicolumn{2}{c}{WER (\%)} & \multicolumn{2}{c}{F1 (\%)} \\ 
    \cmidrule(lr){3-4}
    \cmidrule(lr){5-6}
    & Score & VP & VC & VP & VC \\
    \midrule
    \multicolumn{6}{l}{\textbf{SLUE Benchmark\cite{shon2022slue}}} \\
    \multicolumn{6}{l}{\textbf{Pipeline*:}} \\
    W2V2-L-LL60K & 70.8 & 12.1 & 13.8 & 59.7 & 65.7 \\ 
    W2V2-L-LL60K + LM$^{**}$ & 75.7 & 9.3 & 11.1 & 71.8 & 65.5  \\
    \multicolumn{6}{l}{\textbf{E2E Approach:}} \\
    W2V2-B-LS960 + LM$^{**}$ & 65.7 & 12.3 & 16.1 & 63.4 & 48.1 \\ 
    W2V2-L-LL60K + LM$^{**}$ & 68.1 & 9.3 & 11.1 & 64.8 & 49.8  \\
    \midrule
    \multicolumn{6}{l}{\textbf{Proposed E2E Approach}} \\
    Whisper Small & 68.3 & 14.7 & 16.0 & 59.3 & 60.9 \\
    Whisper Medium & 71.3 & 11.4 & 13.1 & 65.0 & 61.0 \\
    + Classifier & 72.6 & 10.9 & 12.3 & 65.8 & 63.6 \\
    + LLM LoRA finetuning & 74.6 & 10.6 & 11.5 & 68.9 & 65.9 \\
    \bottomrule
  \end{tabular}
  }
  \caption{Performance of the SENSE-ASR models on SLUE-VoxPopli and SLUE-VoxCeleb with SLUE score. (*Pipepline approach used the pre-trained DeBERTa-L model fine-tuned for NER\cite{shon2022slue}. **TED LM with Beam decoding of 500 for ASR and NER task.)}
  \label{tab:slue-table}
\end{table}

\subsection{SLUE Benchmark}

Table \ref{tab:slue-table} presented in the paper provides a comprehensive comparison of the proposed E2E approach against the SLUE Benchmark models on two key tasks: NER and SA. The evaluation is based on the SLUE score defined in section \ref{sec:metrics}, which is an average of the primary metrics for ASR, SA and NER tasks, providing a holistic view of model performance \cite{shon2022slue}. 

The proposed E2E approach achieved significant improvements over the SLUE Benchmark models. When comparing similar-sized models, Whisper Small version achieved a SLUE score of 68.3, outperforming the SLUE W2V2-B-LS960+LM, which scored 65.7. Additionally, Whisper Medium, when compared to the similarly sized SLUE W2V2-L-LL60K+LM, not only surpassed it with a SLUE score of 71.3 versus 68.1, but also demonstrated competitive WER while significantly enhancing F1 scores.

The integration of the classifier further enhanced the SLUE score to 72.6\%, 2\% relative improvement. These results highlight the effectiveness of using a classifier as a loss regularizer within the proposed E2E approach, leading to superior performance across both NER and SA tasks.

Finally, expanding the training parameters with LLM LoRA fine-tuning further enhanced the model's SLUE score to 74.3\%, 2.7\% relative improvement.

Overall comparing our proposed E2E approach best SLUE score result 74.6 against with the benchmark E2E best result we achieved 9.5\% relative improvement. 

Although our proposed models exhibit slightly higher WER scores compared to the benchmark, this is largely due to the benchmark's use of an out-of-domain TED LM with a large beam decoding size of 500. Notably, with an in-domain language model, the benchmark paper reports lower WER scores of 15.2\% and 18.1\% for VP and VC, respectively, using W2V2-B-LS960, and 12.5\% and 13.9\% for VP and VC, respectively, with W2V2-L-LL60K. The integration of advanced techniques such as the classifier and LLM LoRA fine-tuning plays a crucial role in enhancing the model's capability to handle complex speech tasks, thereby setting a new standard for performance in the field.

Despite the benchmark E2E approach (68.1\%) traditionally lagging behind its corresponding pipeline method (75.7\%) on low-resource data, which often delivers superior performance, our proposed E2E approach has successfully bridged this gap (74.6\%). It has achieved results that are significantly better, particularly in terms of the SA F1 score. This demonstrates that our approach not only surpasses the equivalent E2E benchmark method but also achieves performance on par with the pipeline method, highlighting its effectiveness and potential.

\section{Conclusion}
In this paper, we introduced SENSE-ASR, an E2E model that effectively integrates speech and large language models using a novel adapter design to enhance performance on ASR, NER and SA tasks. The proposed model exhibits the capability to capture semantic meanings by effectively mapping speech features to text tokens that are interpretable by LLMs.

Our approach leverages a multi-stage training strategy, starting with ASR data, followed by training on synthetically generated NER labels, and concluding with fine-tuning on low-resource, human-annotated datasets. This strategy not only mitigates the reliance on extensive NER data labeling but also achieves state-of-the-art performance with significantly fewer trainable parameters. Specifically, compared to other similar E2E models, the proposed SENSE-ASR shows a 26\% relative improvement in WER on the Librispeech dataset and achieves a state-of-the-art F1 score of 74.6\% on the SLUE benchmark, a 9.5\% relative improvement comparing to equivalent E2E benchmark model. By achieving benchmark performance on NER and SA and SLUE score, we demonstrate the capability of effectively extending LLM-based ASR models.

\bibliographystyle{IEEEtran}
\bibliography{template}

\end{document}